\documentclass[letterpaper]{article} 
\usepackage[preprint]{aaai2027}  
\usepackage[hyphens]{url}  
\usepackage{graphicx} 
\usepackage{natbib}  
\usepackage{caption} 
\usepackage{algorithm}
\usepackage{algorithmic}

\usepackage{amsmath}
\usepackage{multirow}
\usepackage{arydshln}

\usepackage{newfloat}
\usepackage{listings}
\DeclareCaptionStyle{ruled}{labelfont=normalfont,labelsep=colon,strut=off} 
\floatstyle{ruled}
\newfloat{listing}{tb}{lst}{}
\floatname{listing}{Listing}

\usepackage{booktabs}

\title{GenRubric: Self-Evolving Rubric Generation for Scalable LLM Evaluation}

\author{
    Yifan Chen\textsuperscript{\rm 1}\equalcontrib,
    Haitao Li\textsuperscript{\rm 1}\equalcontrib,
    Qingyao Ai\textsuperscript{\rm 1}\corresponding,
    Fengbin Zhu\textsuperscript{\rm 2},
    Tat-Seng Chua\textsuperscript{\rm 2},\\
    Min Zhang\textsuperscript{\rm 1},
    Yiqun Liu\textsuperscript{\rm 1}
}

\affiliations{
    \textsuperscript{\rm 1}Department of Computer Science and Technology, Tsinghua University\\
    \textsuperscript{\rm 2}National University of Singapore\\[2pt]
    chenyifan26@mails.tsinghua.edu.cn
}

\begin{document}

\maketitle

\begin{abstract}
Large language models are increasingly used as scalable evaluators for open-ended tasks. However, many LLM judges derive query-specific criteria during scoring, leaving the evaluation requirements insufficiently specified and their coverage difficult to audit. Query-specific rubrics make these requirements explicit, but expert-written rubrics are costly to construct, while existing automatic methods typically rely on inference-time refinement or external supervision. We introduce \textsc{GenRubric}, a self-evolving framework that improves rubric generation from unlabeled queries without requiring additional human annotations during self-evolution. Our approach is based on \emph{rubric-induced self-consistency}: independently sampled rubrics for the same query provide partial views of its latent evaluation requirements, and a comprehensive rubric should induce a response that generalizes across these complementary evaluation views. We implement this principle through reinforcement learning, combining a cross-rubric comprehensiveness signal with group-level and criterion-level rewards for rubric quality. We train \textsc{GenRubric} models at 4B, 8B, and 14B scales across multiple domains. Experiments on human-annotated rubric benchmarks show that self-evolution improves the agreement between evaluations induced by generated rubrics and those induced by expert-written rubrics. The improvements further generalize to held-out domains, demonstrating the potential of self-evolving rubric generation for scalable and query-specific LLM evaluation. Code and models are publicly available at \url{https://github.com/foggpoy/GenRubric}.
\end{abstract}


\section{Introduction}

Large language models (LLMs) are increasingly deployed to support real-world decision making. 
For tasks with objectively verifiable outcomes, model outputs can typically be evaluated against reference answers or using programmatic verifiers~\cite{hendrycks2021math,chen2021codex}. By contrast, many open-ended tasks in domains such as medicine and law admit multiple valid responses, whose quality depends on factual correctness, information coverage, risk awareness, practical utility, and other query-specific considerations~\cite{chang2024survey,arora2025healthbench,akyurek2025prbench}.
LLM-as-a-Judge provides a scalable approach to such evaluation ~\cite{liu2023geval,zheng2023judging}. However, its operational evaluation criteria are often instantiated during scoring rather than explicitly specified
beforehand, coupling criterion construction with the final
judgment. This can blur the boundaries between distinct requirements, leave their evaluation boundaries underspecified, and make criterion coverage difficult to audit. It can also make the resulting evaluations less trustworthy, as it is difficult to guarantee that consistent criteria are applied to all candidate responses for the same query.

Query-specific rubrics address this need by decomposing response quality into fine-grained and checkable criteria. Expert-written rubrics have enabled structured evaluation across diverse open-ended tasks ~\cite{arora2025healthbench,akyurek2025prbench,
sharma2025researchrubrics,chen2026lexrubric}, while rubric-conditioned evaluators have demonstrated the value of customized criteria for fine-grained assessment ~\cite{kim2024prometheus}. Rubrics can also provide criterion-level signals for reward modeling, making the resulting supervision more structured and interpretable~\cite{gunjal2025rar}.

Yet constructing a dedicated rubric for each query requires substantial domain expertise and annotation effort, making it difficult to scale to new tasks and queries
~\cite{arora2025healthbench,sharma2025researchrubrics}. Generating rubrics with LLMs is a natural alternative, but off-the-shelf LLM-generated rubrics may be insufficiently aligned with human-authored criteria, particularly in factual and knowledge-intensive settings ~\cite{dhole2026rubricrag,siro2026learning}. These limitations have motivated the design of dedicated rubric generators.

Recent studies on rubric generators commonly rely on pre-collected annotated supervision, most often preference labels over model responses ~\cite{liu2025openrubrics,lv2026queryspecific,
kawabata2026c2,xu2026rubricarm}. Such supervision is useful for identifying criteria that distinguish the observed responses, but the resulting rubrics inherit the coverage of the annotations and response distributions from which they are learned. They may therefore capture criteria that are discriminative for the available examples without
fully representing the evaluation requirements of the query itself, while collecting comparable supervision for new domains remains costly. This motivates our question: \emph{Can a rubric generator improve solely from unlabeled queries and its own generated rubrics, without requiring pre-collected supervision for each query?} 

We observe that independently generated rubrics for the same query provide multiple noisy and partial views of its latent evaluation requirements. Inspired by majority aggregation in collective decision making, where agreement across noisy individual judgments can reveal their shared signal, we seek consensus among these independently sampled views. Directly voting over rubric text is inappropriate because valid rubrics may capture complementary rather than identical requirements. We therefore define consensus through their behavioral consequences: a comprehensive rubric should induce a response that is also highly rated by other independently generated rubrics for the same query. A narrow rubric, in contrast, is more likely to induce a response tailored to its own limited criteria. This observation enables rubric quality to be estimated implicitly without additional supervision.

Based on this insight, we propose \textsc{GenRubric}, a self-evolving reinforcement learning framework that improves the rubric generation with unlabeled queries. Its training objective combines a sequence-level cross-rubric comprehensiveness reward, a group-level reward that encourages
complete rubric groups to distinguish responses of different quality, and criterion-level rewards for discriminativeness and non-redundancy that provide fine-grained feedback to individual rubric items. The self-evolution stage requires no additional human annotations. We train \textsc{GenRubric} models at 4B, 8B, and 14B scales across multiple domains. Experiments on human-annotated rubric benchmarks show that our method improves the agreement between evaluations induced by generated rubrics and those induced by expert-written rubrics. These improvements also extend to held-out domains not observed during training, demonstrating the cross-domain generalization of the learned rubric-generation capability.

\section{Related Work}

\textbf{Training-Free Rubric Generation.}
Training-free methods use fixed LLMs to construct or refine rubrics at inference time. CARMO and GER-Eval dynamically elicit context- or task-specific criteria, while RubricHub, RRD, and RubricRAG improve rubric quality through coarse-to-fine synthesis, recursive refinement, or retrieval augmentation~\cite{gupta2025carmo,siro2026learning,li2026rubrichub,shen2026rethinking,dhole2026rubricrag}. Wang and Blanco further generate dataset- and instance-specific rubrics without human annotations or reference answers~\cite{wang2026dynamicrubrics}. These methods make rubric construction more scalable than expert authoring, but their improvements generally rely on additional inference-time generation, refinement, or auxiliary context, while the underlying generator remains fixed. In contrast, \textsc{GenRubric} learns a reusable query-to-rubric capability, incorporating rubric improvement into model parameters for direct generation on new queries.

\textbf{Trainable Rubric Generators.}
Most trainable approaches derive supervision from response preferences. OpenRubrics, the DeepResearch rubric generator of Lv et al., C2, and RubricARM use pairwise or binary preferences to train rubric generators, either independently or jointly with rubric-conditioned judges~\cite{liu2025openrubrics,lv2026queryspecific,kawabata2026c2,xu2026rubricarm}. Such supervision is effective for discovering criteria that distinguish observed responses, but is constrained by the available response and preference distributions, while collecting reliable preferences remains costly in new domains. Recent annotation-free methods instead refine generators using rubric-level preferences from a strong meta-judge or jointly co-evolve rubrics with a downstream policy~\cite{wang2026dynamicrubrics,ding2026evorubrics}. \textsc{GenRubric} differs by evaluating rubrics through the cross-rubric generalization of their induced responses. This rubric-induced self-consistency provides group- and criterion-level learning signals from unlabeled queries, without direct rubric-level preference labels or coupling the generator to downstream policy optimization.

\section{Method}

Given an open-ended query $x$, our goal is to generate a query-specific rubric group
\begin{equation}
R =
\left[
(c_1,w_1),
\ldots,
(c_K,w_K)
\right],
\label{eq:rubric_definition}
\end{equation}
where each rubric item $(c_k,w_k)$ consists of an evaluation criterion $c_k$ and its associated importance weight $w_k$. Positive weights specify desirable properties of a response, whereas negative weights specify observable errors that should be penalized. Each criterion should describe a specific and verifiable requirement, while the complete rubric group should comprehensively cover the query and distinguish responses of different quality. We learn a rubric generator $\pi_\theta(R\mid x)$ that produces the rubric group in a structured format.

The central challenge is that rubric quality has no directly verifiable scalar target. A rubric is useful only when its criteria can effectively guide response generation and distinguish responses of different quality. We therefore construct a closed training loop in which candidate rubric
groups are first used to induce responses and are then evaluated through their interactions with other rubric groups sampled for the same query.

\begin{figure*}[t]
    \centering
    \includegraphics[width=\textwidth]{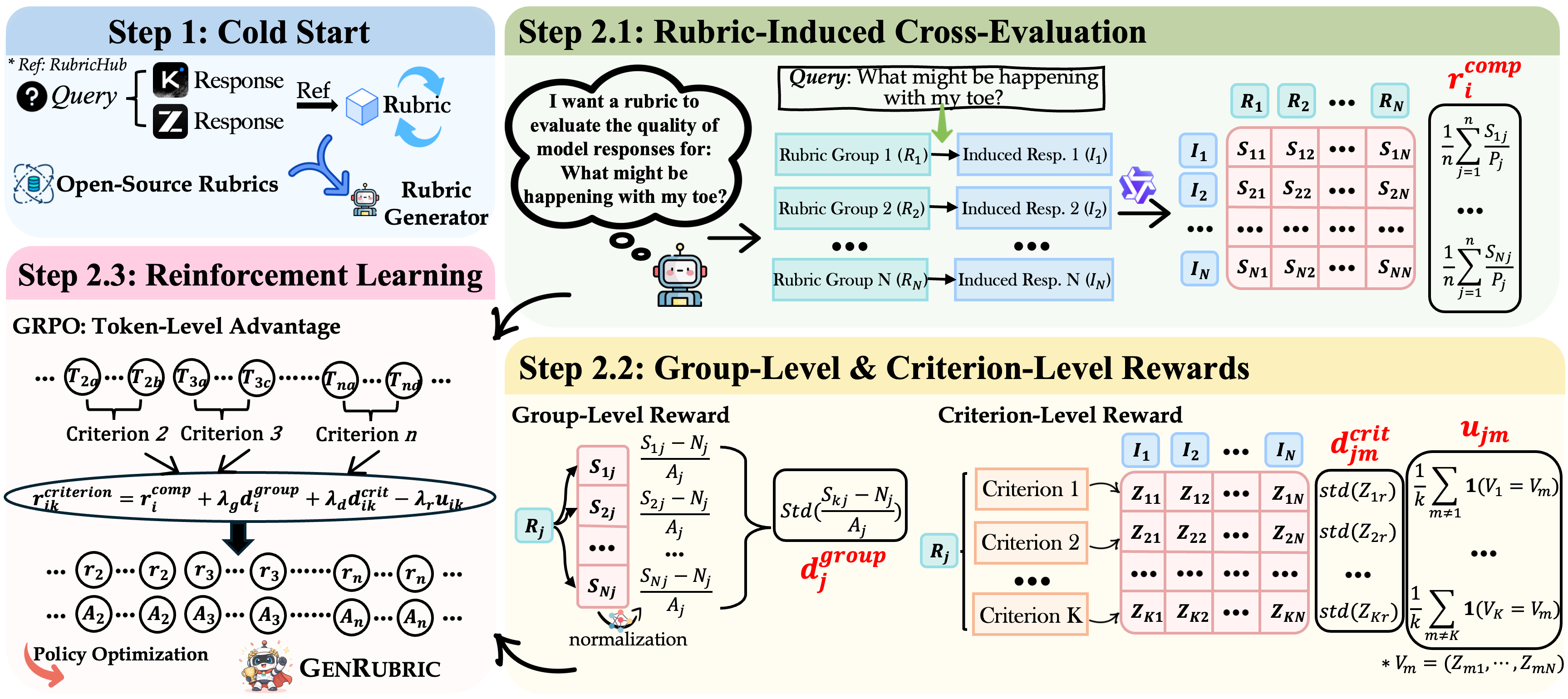}
    \caption{Overview of \textsc{GenRubric} framework.}
    \label{fig:framework}
\end{figure*}

As illustrated in Figure~\ref{fig:framework}, \textsc{GenRubric} starts from a cold-start rubric generator obtained through supervised fine-tuning. During self-evolution, the policy samples $n$ candidate rubric groups for each unlabeled query. After excluding malformed outputs, we reindex the $m$ valid candidates as $\{R_i\}_{i=1}^{m}$. A fixed LLM is prompted to serve in two roles: a response generator $\mathcal{G}$ and a rubric judge $\mathcal{J}$. The resulting cross-evaluation process provides three complementary learning signals:
\begin{enumerate}
    \item \emph{Cross-rubric comprehensiveness}: whether a rubric group covers requirements from multiple evaluation perspectives;
    \item \emph{Group-level discrimination}: whether a complete rubric group distinguishes responses of different quality;
    \item \emph{Criterion-level discrimination and redundancy}: the evaluation utility of individual rubric items.
\end{enumerate}
Invalid rubric outputs receive zero reward and do not participate in cross-evaluation.

\subsection{Rubric-Induced Cross-Evaluation}

For each valid candidate rubric group $R_i$, we condition the response generator on both the query and the rubric to produce a rubric-induced response $y_i$. A rubric explicitly specifies the properties that a response should satisfy or avoid. When a rubric accurately captures the evaluation requirements of the query, conditioning generation on that rubric should help the resulting response cover these requirements. Consequently, the quality of the induced response provides behavioral evidence about the quality of the rubric.

We then use every valid rubric group $R_j$ to evaluate every induced response $y_i$. Suppose that
\begin{equation}
R_j =
\left[
(c_{j1},w_{j1}),
\ldots,
(c_{jK_j},w_{jK_j})
\right].
\label{eq:candidate_rubric}
\end{equation}
Let
\begin{equation}
z_{jik}
=
\mathcal{J}(x,y_i,c_{jk})
\in \{0,1\}
\label{eq:criterion_judgment}
\end{equation}
denote whether the fixed rubric judge $\mathcal{J}$ determines that response $y_i$ exhibits the property described by the $k$-th criterion of rubric group $R_j$. For a positive criterion, $z_{jik}=1$ indicates that the requirement is satisfied. For a negative criterion, it indicates that the specified undesirable behavior is present. The score assigned by rubric group $R_j$ to response $y_i$ is
\begin{equation}
s_{ij}
=
\sum_{k=1}^{K_j}
w_{jk}z_{jik}.
\label{eq:cross_rubric_score}
\end{equation}

Different rubric groups may contain different numbers of criteria and different total weights. We define the maximum attainable score $P_j$, the minimum attainable
score $N_j$, and the absolute score range $A_j$ as
\begin{equation}
\begin{aligned}
P_j &= \sum_{k:w_{jk}>0} w_{jk}, \\
N_j &= \sum_{k:w_{jk}<0} w_{jk}, \\
A_j &= \sum_{k=1}^{K_j} |w_{jk}|.
\end{aligned}
\label{eq:rubric_score_ranges}
\end{equation}


Our principal reward follows the rubric-induced self-consistency
principle. Independently sampled rubric groups can be viewed as partial and noisy observations of the latent evaluation requirements of the same query. If $R_i$ is comprehensive, the response induced by it should not only satisfy its own criteria, but also satisfy complementary requirements captured by other sampled rubric groups.

Since different rubric groups have different maximum attainable positive scores, we normalize each cross-rubric score by the corresponding $P_j$ before aggregation. The cross-rubric comprehensiveness reward is defined as
\begin{equation}
r_i^{\mathrm{comp}}
=
\frac{1}{m}
\sum_{j=1}^{m}
\frac{s_{ij}}{P_j}.
\label{eq:comprehensiveness_reward}
\end{equation}
A high value indicates that the response induced by $R_i$ performs well across multiple independently sampled evaluation views, providing a behavioral proxy for the comprehensiveness of $R_i$.

\subsection{Group-Level and Criterion-Level Rewards}

The comprehensiveness reward measures whether a rubric group elicits a response that broadly satisfies the query requirements. Since rubrics are also used to compare candidate responses, they must additionally assign meaningfully different scores to responses of different quality. We therefore introduce a group-level discrimination reward.

The score assigned by rubric group $R_j$ to response $y_i$ is normalized as
\begin{equation}
\widetilde{s}_{ij}
=
\frac{s_{ij}-N_j}{A_j}
\in [0,1].
\label{eq:normalized_rubric_score}
\end{equation}
The group-level discrimination of rubric group $R_j$ is defined as the standard deviation of its normalized scores over all induced responses:
\begin{equation}
d_j^{\mathrm{group}}
=
\mathrm{Std}_{i}
\left(
\widetilde{s}_{ij}
\right).
\label{eq:group_discrimination}
\end{equation}
A rubric group that assigns similar scores to all responses receives a low discrimination reward, whereas a rubric group that exposes meaningful quality differences receives a higher value. The sequence-level reward of
$R_i$ is
\begin{equation}
r_i^{\mathrm{seq}}
=
r_i^{\mathrm{comp}}
+
\lambda_g d_i^{\mathrm{group}}.
\label{eq:sequence_reward}
\end{equation}

We further derive criterion-level signals from the binary judgment
patterns. For criterion $c_{ik}$, its discrimination is defined as
\begin{equation}
d_{ik}^{\mathrm{crit}}
=
\mathrm{Std}_{r}
\left(
z_{irk}
\right),
\label{eq:criterion_discrimination}
\end{equation}
where the standard deviation is computed over all $m$ induced responses. A criterion that is activated by every response or by no response provides no distinction among the sampled responses. In contrast, a criterion whose judgments vary across responses provides a stronger signal for response comparison.

To measure functional redundancy, we represent each criterion using its activation fingerprint over all induced responses:
\begin{equation}
\mathbf{v}_{ik}
=
\left(
z_{i1k},
\ldots,
z_{imk}
\right).
\label{eq:criterion_fingerprint}
\end{equation}
The redundancy score of criterion $c_{ik}$ is
\begin{equation}
u_{ik}
=
\frac{1}{K_i}
\sum_{k'\neq k}
\mathbf{1}
\left(
\mathbf{v}_{ik}
=
\mathbf{v}_{ik'}
\right),
\label{eq:criterion_redundancy}
\end{equation}
where $k'$ ranges over all criteria in $R_i$. Two criteria with identical activation fingerprints behave equivalently on the sampled responses and are therefore treated as functionally redundant.

The final reward assigned to rubric item $(c_{ik},w_{ik})$ is
\begin{equation}
r_{ik}^{\mathrm{item}}
=
r_i^{\mathrm{seq}}
+
\lambda_d d_{ik}^{\mathrm{crit}}
-
\lambda_r u_{ik}.
\label{eq:item_reward}
\end{equation}
These rewards jointly encourage the generated rubric groups to
comprehensively cover the query requirements, effectively distinguish responses, and reduce functionally repetitive criteria.

\subsection{Criterion-Level Credit Assignment}

In common outcome-supervised LLM reinforcement learning settings based on PPO or GRPO, each generated sequence is assigned a single scalar reward~\cite{schulman2017proximal,shao2024deepseekmath}. When an entire rubric group is generated as one sequence, such coarse-grained feedback makes it difficult to identify which individual rubric items contribute to the reward. We instead locate the token span $\mathcal{T}_{ik}$ corresponding to the complete serialized rubric item $(c_{ik},w_{ik})$, including both its criterion and point value.

The token-level reward is defined as
\begin{equation}
r_{i,t}
=
\left\{
\begin{array}{ll}
r_{ik}^{\mathrm{item}},
&
t\in\mathcal{T}_{ik},
\\[2pt]
r_i^{\mathrm{seq}},
&
\mbox{otherwise}.
\end{array}
\right.
\label{eq:token_reward}
\end{equation}
Thus, all tokens belonging to a complete rubric item receive its item-level reward, whereas structural tokens outside rubric items, such as the surrounding JSON array syntax, receive the sequence-level reward.

We optimize these rewards using a token-level variant of Group Relative Policy Optimization ~\cite{shao2024deepseekmath}. Let $\mu_{i,t}\in\{0,1\}$ indicate whether token $t$ is a valid response token in the $i$-th rollout. For all rollouts sampled from the same query, we compute a group baseline as the mean reward over valid response tokens:
\begin{equation}
b_x
=
\frac{
\displaystyle
\sum_{i=1}^{m}\sum_t
r_{i,t}\mu_{i,t}
}{
\displaystyle
\sum_{i=1}^{m}\sum_t
\mu_{i,t}
}.
\label{eq:token_group_baseline}
\end{equation}
The advantage of each token is
\begin{equation}
A_{i,t}
=
\left(
r_{i,t}-b_x
\right)
\mu_{i,t}.
\label{eq:token_advantage}
\end{equation}
We do not normalize the advantages by the group standard deviation, since the absolute reward differences within a rubric group encode the relative quality of its individual criteria. The resulting token-level advantages are used in the standard clipped policy objective.

This formulation provides two levels of credit assignment:
\begin{itemize}
    \item \emph{Across different rollouts,} a rubric group with higher overall quality receives a higher advantage;
    \item \emph{Within the same rollout,} a more discriminative and less redundant criterion receives a stronger gradient signal.
\end{itemize}

The response-generation and cross-evaluation procedures are used only during self-evolution. At inference time, \textsc{GenRubric} directly maps a new query to its query-specific rubric group.

\section{Experiments}
\label{sec:experiments}

\begin{table*}[t]
\centering
\begin{tabular}{ll|cccc:cc}
\hline
Category & Model & $\rho$ & $\tau_b$ & Top-1 & Pairwise
& Rubric Len. & Crit. Len. \\
\hline
\multirow{7}{*}{\textbf{\textit{General LLMs}}}
& Kimi-K2.5             & 0.3670 & 0.3038 & 0.1449 & 0.5191
                          & 348.2 & 31.73 \\
& GPT-5.2               & 0.4343 & 0.3521 & 0.2182 & 0.5798
                          & 646.6 & 52.44 \\
& Qwen3-Max             & 0.3098 & 0.2589 & 0.0688 & 0.4274
                          & 302.9 & 35.47 \\
& Qwen3.5-397B-A17B     & 0.2894 & 0.2436 & 0.0659 & 0.4351
                          & 213.7 & 23.17 \\
& GLM-5                 & 0.3092 & 0.2603 & 0.0747 & 0.4213
                          & 227.6 & 29.21 \\
& DeepSeek-V3.2         & 0.2753 & 0.2311 & 0.0600 & 0.4166
                          & 299.3 & 34.72 \\
& Claude Sonnet 4.6     & 0.4059 & 0.3345 & 0.1391 & 0.5383
                          & 510.5 & 38.24 \\
\hline
\multirow{5}{*}{\textbf{\textit{Rubric Generators}}}
& DeepResearch Gen.     & 0.2724 & 0.2348 & 0.0484 & 0.3615
                          & 325.9 & 28.08 \\
& RubricARM-8B          & 0.2441 & 0.2120 & 0.0367 & 0.3129
                          & 144.2 & 15.10 \\
& RubricARROW-8B        & 0.2210 & 0.1950 & 0.0264 & 0.2687
                          & 108.9 & 17.24 \\
& RubricRM-4B           & 0.1798 & 0.1552 & 0.0440 & 0.3291
                          & 183.8 & 14.94 \\
& RubricRM-8B           & 0.2606 & 0.2272 & 0.0601 & 0.3532
                          & 187.5 & 15.09 \\
\hline
\multirow{6}{*}{\textbf{\textit{Qwen3 Baselines}}}
& Qwen3-4B-Base         & 0.1508 & 0.1242 & 0.1036 & 0.4008
                          & 236.6 & 17.52 \\
& Qwen3-4B              & 0.1493 & 0.1281 & 0.0660 & 0.3436
                          & 190.3 & 19.58 \\
& Qwen3-8B-Base         & 0.1695 & 0.1408 & 0.1113 & 0.4032
                          & 236.0 & 17.03 \\
& Qwen3-8B              & 0.2061 & 0.1724 & 0.0880 & 0.3997
                          & 232.4 & 20.92 \\
& Qwen3-14B-Base        & 0.1603 & 0.1327 & 0.1054 & 0.4151
                          & 286.2 & 18.47 \\
& Qwen3-14B             & 0.1777 & 0.1453 & 0.0850 & 0.4130
                          & 266.4 & 24.18 \\
\hline
\multirow{3}{*}{\textbf{\textit{Ours}}}
& GenRubric-4B          & 0.4272 & 0.3404 & 0.3205 & 0.5943
                          & 1,199.0 & 27.29 \\
& GenRubric-8B          & \underline{0.4525} & \underline{0.3638}
                          & \textbf{0.3761} & \underline{0.6089}
                          & 1,253.0 & 32.37 \\
& GenRubric-14B         & \textbf{0.4568} & \textbf{0.3698}
                          & \underline{0.3628} & \textbf{0.6091}
                          & 1,265.9 & 33.36 \\
\hline
\end{tabular}
\caption{
Main results on the 700-query expert-rubric evaluation set. All reported values are averaged over the 700 queries. Rubric Len.\ and Crit.\ Len.\ are the average token lengths of a complete serialized rubric group and an individual criterion, respectively. Best and second-best results in each column are bolded and underlined based on unrounded values.
}
\label{tab:main_results}
\end{table*}

\subsection{Experimental Setup}

\noindent\textbf{Cold-start.}
We first perform supervised fine-tuning to initialize the
rubric-generation capability. We use the 180K examples released by RubricHub~\cite{li2026rubrichub}. We additionally synthesize 27K query--rubric instances covering code, finance, legal, and deep research tasks by adapting the RubricHub generation framework. Specifically, we extend the original pipeline to generate conservative negative criteria that penalize explicit and observable errors. The synthesis queries are drawn in part from OpenCodeInstruct, the code-generation subset of LiveCodeBench, O-Researcher-RL-Dataset, FinanceReasoning, and Fino1, together with additional domain queries~\cite{
ahmad2025opencodeinstruct,jain2024livecodebench,
yao2026oresearcher,tang2025financereasoning,qian2025fino1}.

\noindent\textbf{Training.}
We train three versions of \textsc{GenRubric} initialized from
Qwen3-4B-Base, Qwen3-8B-Base, and Qwen3-14B-Base~\cite{yang2025qwen3}. Details of the training data are provided in the Technical Supplement. During reinforcement learning with the veRL framework~\cite{sheng2024hybridflow}, we sample eight rubric groups per query and use a batch size of 64. The learning rate is $3\times10^{-6}$, and the maximum prompt and response lengths are 8,192 and 2,048 tokens, respectively. The coefficients of group-level discrimination, criterion-level discrimination, and redundancy are all set to $0.1$. Training is conducted on two NVIDIA H800 GPUs.

\noindent\textbf{Evaluation protocol.}
We evaluate 700 queries with expert-written rubrics across healthcare, finance, law, and DeepResearch: 300 sampled HealthBench queries, 150 queries each from the PRBench Finance and Legal Hard subsets, and 100 ResearchRubrics queries, all sampled with seed $42$~\cite{arora2025healthbench,akyurek2025prbench,
sharma2025researchrubrics}.

For each query, we pre-generate responses from Qwen3-4B, Qwen3-14B, Qwen3-32B, Qwen3-235B-A22B, Kimi-K2.5, Claude Sonnet 4.6, DeepSeek-V3.2, and GPT-5.2~\cite{kimiteam2026kimik25visualagentic,deepseek2025v32}. Because generated and expert rubrics may have different score ranges, we compare their induced response rankings using Spearman's $\rho$ and tie-aware Kendall's $\tau_b$ for overall ranking agreement, Top-1 consistency for whether the generated-rubric top response belongs to the expert-rubric top set, and pairwise accuracy for the proportion of consistently ordered response pairs. Formal definitions are provided in the Technical Supplement. Qwen3.5-27B with temperature $0.0$ serves as the rubric judge, determining whether a response satisfies each criterion. Judge-replacement results in the Technical Supplement show that the main conclusions remain stable across alternative judges.

\noindent\textbf{Baselines.}
We compare against seven general-purpose models: Kimi-K2.5, GPT-5.2, Qwen3-Max, Qwen3.5-397B-A17B, GLM-5, DeepSeek-V3.2, and Claude Sonnet 4.6. Specialized rubric-generation baselines include RubricRM-4B, RubricRM-8B, RubricARM-8B, RubricARROW-8B, and the DeepResearch-specific rubric generator released by Lv et al.~\cite{
liu2025openrubrics,xu2026rubricarm,jiang2026rubricarrow,
lv2026queryspecific}.

\subsection{Main Results}

\begin{figure*}[t]
\centering
\includegraphics[width=\textwidth]{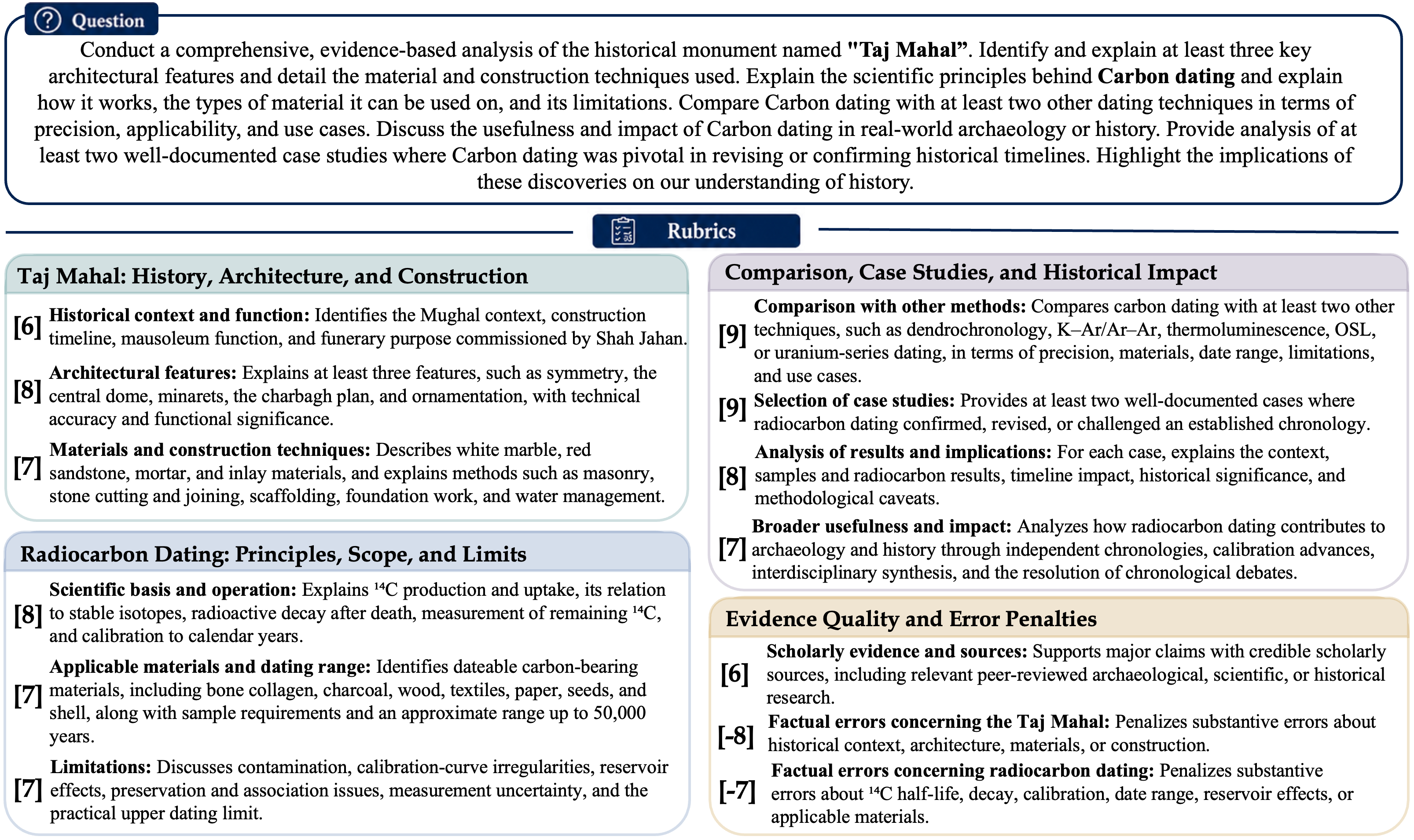}
\caption{An example of rubrics generated by \textsc{GenRubric}.}
\label{fig:case_study}
\end{figure*}

Table~\ref{tab:main_results} shows that \textsc{GenRubric} overall achieves stronger agreement with expert rubrics than existing specialized rubric generators and general-purpose LLMs. The improvements hold across all three model sizes, demonstrating that self-evolution is effective rather than being tied to a particular parameter scale. Performance on the rank-correlation and pairwise metrics generally improves with model size, with the largest gains occurring from 4B to 8B and further improvements observed at 14B. Meanwhile, \textsc{GenRubric}-4B is already competitive with the strongest general-purpose models and surpasses all of them on Top-1 consistency and pairwise accuracy. These results suggest that model scaling provides additional benefits, while the effectiveness of \textsc{GenRubric} primarily stems from the proposed self-evolution framework. Figure~\ref{fig:case_study} shows a representative example generated by \textsc{GenRubric}. Notably, on ResearchRubrics, the DeepResearch-specific rubric generator achieves a Spearman correlation of $0.322$, whereas \textsc{GenRubric}-4B already reaches $0.603$ despite being trained across multiple domains. This margin shows that \textsc{GenRubric} achieves cross-domain generality without sacrificing domain-specific effectiveness, outperforming even a generator tailored to DeepResearch.

Stronger systems generally produce more detailed rubrics.
\textsc{GenRubric} generates longer complete rubric groups while retaining moderate criterion lengths, suggesting that the additional tokens are primarily used to cover more evaluation requirements. The results also indicate a positive association between rubric coverage and evaluation agreement, although length alone is not sufficient to guarantee strong performance.


\subsection{Ablation Studies}

\noindent\textbf{Training stages.}
Table~\ref{tab:stage_ablation} examines the effects of cold-start SFT and self-evolutionary RL. At every model size, SFT substantially improves over the corresponding base model, and RL further improves all four metrics. The two stages serve complementary roles. \emph{SFT equips the base model with
structured rubric-generation, format-following, and basic task-solving capabilities, whereas RL further optimizes the evaluation utility of the generated rubrics through rubric-induced self-consistency.}

Direct RL remains effective without cold-start supervision, providing direct evidence that the self-evolution signal itself is informative. Starting directly from Qwen3-Base, both the 8B and 14B models improve on the main rank-correlation metrics, although the gains are less uniform for
14B. One possible explanation is that direct RL uses the same training budget and optimization setting for base models that have not undergone instruction post-training. Without SFT initialization, the larger model may require more data or optimization steps to reliably follow complex instructions and exploit the fine-grained reward signals. Cold-start SFT
provides instruction-following capabilities, relevant task knowledge, and a stronger initialization for rubric generation. SFT therefore helps the model exploit the RL signal more effectively, but is not necessary for the signal itself to be informative. Due to computational constraints, direct base-to-RL training is evaluated only at 8B and 14B.

\begin{table}[t]
\centering
\begin{tabular}{l|cccc}
\hline
Training stage & $\rho$ & $\tau_b$ & Top-1 & Pair. \\
\hline
\textit{Qwen3-4B-Base}
                    & 0.151 & 0.124 & 0.104 & 0.401 \\
\quad w/ SFT        & 0.286 & 0.227 & 0.264 & 0.544 \\
\quad \textbf{w/ SFT + RL}
                    & \textbf{0.427} & \textbf{0.340}
                    & \textbf{0.321} & \textbf{0.594} \\
\hline
\textit{Qwen3-8B-Base}
                    & 0.169 & 0.141 & 0.111 & 0.403 \\
\quad w/ SFT        & 0.324 & 0.254 & 0.274 & 0.557 \\
\quad \textbf{w/ SFT + RL}
                    & \textbf{0.452} & \textbf{0.364}
                    & \textbf{0.376} & \textbf{0.609} \\
\hline
\textit{Qwen3-14B-Base}
                    & 0.160 & 0.133 & 0.105 & 0.415 \\
\quad w/ SFT        & 0.383 & 0.308 & 0.323 & 0.585 \\
\quad \textbf{w/ SFT + RL}
                    & \textbf{0.457} & \textbf{0.370}
                    & \textbf{0.363} & \textbf{0.609} \\
\hline
\textit{Direct RL} \\
\quad 8B Base w/ RL       & 0.281 & 0.233 & 0.206 & 0.489 \\
\quad 14B Base w/ RL      & 0.256 & 0.218 & 0.111 & 0.393 \\
\hline
\end{tabular}
\caption{
Ablation of training stages.}
\label{tab:stage_ablation}
\end{table}

\noindent\textbf{Reward components.}
Table~\ref{tab:reward_ablation} identifies cross-rubric behavioral consensus as the primary learning signal in \textsc{GenRubric}. Starting from the same SFT initialization, optimizing only $r^{\mathrm{comp}}$ raises Spearman correlation from $0.324$ to $0.438$ and produces substantial gains across all metrics. This supports our central insight that rubric quality can be inferred from the cross-rubric generalization of rubric-induced responses.
The auxiliary rewards yield comparatively smaller and non-monotonic intermediate changes, suggesting that the main improvement stems from this core signal rather than from explicitly optimizing discrimination and redundancy. Nevertheless, these rewards remain collectively beneficial:
combining all four components achieves the best result on every metric, indicating that they provide complementary refinement to the principal comprehensiveness reward.

\begin{table}[t]
\centering
\begin{tabular}{l|cccc}
\hline
RL objective & $\rho$ & $\tau_b$ & Top-1 & Pair. \\
\hline
No RL (SFT)                                  & 0.324 & 0.254 & 0.274 & 0.557 \\
$r^{\mathrm{comp}}$                          & 0.438 & 0.353 & 0.373 & 0.602 \\
$r^{\mathrm{comp}}+d^{\mathrm{group}}$        & 0.444 & 0.357 & 0.348 & 0.606 \\
$r^{\mathrm{comp}}+d^{\mathrm{group}}
+d^{\mathrm{crit}}$                          & 0.443 & 0.356 & 0.344 & 0.604 \\
\textbf{Full objective}                      & \textbf{0.452}
                                               & \textbf{0.364}
                                               & \textbf{0.376}
                                               & \textbf{0.609} \\
\hline
\end{tabular}
\caption{
Cumulative reward ablation on \textsc{GenRubric}-8B. All RL runs start from the same SFT model; the full objective additionally includes the redundancy penalty $-u$.
}
\label{tab:reward_ablation}
\end{table}

\noindent\textbf{Reward coefficients.}
Table~\ref{tab:coefficient_ablation} examines the sensitivity to the auxiliary reward coefficients. We conduct this analysis at the 8B scale, varying one auxiliary coefficient at a time while keeping the other two fixed at $0.1$. All tested RL configurations remain substantially stronger than the SFT initialization, showing that the effectiveness of self-evolution does not depend on a narrow coefficient range. The default setting of $0.1$ achieves the best performance. This suggests that the auxiliary objectives are most effective
when they refine the principal cross-rubric comprehensiveness signal rather than dominate optimization. Excessive emphasis on discrimination or redundancy reduction may disrupt the balance between these local rubric properties and overall requirement coverage.

\begin{table}[t]
\centering
\begin{tabular}{l|cccc}
\hline
Configuration & $\rho$ & $\tau_b$ & Top-1 & Pair. \\
\hline
SFT only                 & 0.324 & 0.254 & 0.274 & 0.557 \\
\textbf{Default ($0.1$)} & \textbf{0.452} & \textbf{0.364}
                          & \textbf{0.376} & \textbf{0.609} \\
$\lambda_d=0.3$          & 0.424 & 0.340 & 0.327 & 0.596 \\
$\lambda_d=0.5$          & 0.409 & 0.327 & 0.319 & 0.587 \\
$\lambda_r=0.3$          & 0.428 & 0.341 & 0.317 & 0.596 \\
$\lambda_r=0.5$          & 0.418 & 0.336 & 0.335 & 0.597 \\
$\lambda_g=0.3$          & 0.417 & 0.333 & 0.337 & 0.596 \\
$\lambda_g=0.5$          & 0.424 & 0.340 & 0.333 & 0.598 \\
\hline
\end{tabular}
\caption{
Coefficient analysis of \textsc{GenRubric}-8B.}
\label{tab:coefficient_ablation}
\end{table}

\subsection{Analysis}

\noindent\textbf{Cross-domain generalization.}
We evaluate the non-finance domains of ProfBench, which are absent from both cold-start SFT and self-evolution training~\cite{wang2025profbench}. We use the three model responses provided by ProfBench and compare how generated
and expert rubrics rank them. Table~\ref{tab:ood_generalization} reports the average of two evaluation runs. All three \textsc{GenRubric} models outperform their corresponding base models on every metric. \textsc{GenRubric}-14B obtains the highest rank correlations and pairwise accuracy, while \textsc{GenRubric}-8B achieves the highest Top-1 consistency. These results demonstrate that the learned rubric-generation capability transfers to unseen professional domains.

\begin{table}[t]
\centering
\begin{tabular}{l|cccc}
\hline
Model & $\rho$ & $\tau_b$ & Top-1 & Pair. \\
\hline
Kimi-K2.5          &  0.121 &  0.103 & 0.283 & 0.450 \\
GPT-5.2            &  0.193 &  0.179 & 0.283 & 0.494 \\
Qwen3-Max          &  0.035 &  0.034 & 0.207 & 0.356 \\
GLM-5              &  0.041 &  0.037 & 0.300 & 0.411 \\
DeepSeek-V3.2      & -0.098 & -0.092 & 0.140 & 0.287 \\
Claude Sonnet 4.6  &  0.147 &  0.136 & 0.283 & 0.467 \\
\hline
Qwen3-4B-Base      & 0.041 & 0.040 & 0.161 & 0.339 \\
GenRubric-4B       & 0.198 & 0.166 & 0.383 & 0.494 \\
Qwen3-8B-Base      & 0.122 & 0.121 & 0.250 & 0.387 \\
GenRubric-8B       & \underline{0.210} & \underline{0.185}
                    & \textbf{0.400} & \underline{0.517} \\
Qwen3-14B-Base     & 0.178 & 0.132 & 0.286 & 0.446 \\
GenRubric-14B      & \textbf{0.331} & \textbf{0.314}
                    & \underline{0.383} & \textbf{0.556} \\
\hline
\end{tabular}
\caption{
Cross-domain evaluation on the non-finance domains of ProfBench.}
\label{tab:ood_generalization}
\end{table}

\noindent\textbf{Discriminating strong responses.}
A useful rubric group should distinguish not only clearly weak and strong responses, but also responses near the high-quality frontier. For each query in the evaluation set, we retain the three responses ranked highest by its expert rubric and recompute agreement within this restricted set. As shown in Table~\ref{tab:top3_analysis}, \textsc{GenRubric}-8B achieves
the best result on all four metrics. All \textsc{GenRubric} variants substantially improve over their corresponding base models, indicating that their criteria capture sufficiently fine-grained requirements to differentiate among already strong responses.

\begin{table}[t]
\centering
\begin{tabular}{l|cccc}
\hline
Model & $\rho$ & $\tau_b$ & Top-1 & Pair. \\
\hline
Kimi-K2.5          & 0.138 & 0.135 & 0.351 & 0.400 \\
GPT-5.2            & 0.171 & 0.155 & 0.425 & 0.439 \\
Qwen3-Max          & 0.109 & 0.101 & 0.261 & 0.344 \\
Qwen3.5-397B-A17B  & 0.104 & 0.098 & 0.268 & 0.373 \\
GLM-5              & 0.037 & 0.036 & 0.252 & 0.341 \\
DeepSeek-V3.2      & 0.070 & 0.061 & 0.253 & 0.358 \\
Claude Sonnet 4.6  & 0.159 & 0.144 & 0.359 & 0.418 \\
\hline
Qwen3-4B-Base      & 0.058 & 0.058 & 0.322 & 0.369 \\
GenRubric-4B       & 0.150 & 0.135 & 0.483 & 0.452 \\
Qwen3-8B-Base      & 0.024 & 0.022 & 0.315 & 0.352 \\
GenRubric-8B       & \textbf{0.209} & \textbf{0.195}
                    & \textbf{0.552} & \textbf{0.483} \\
Qwen3-14B-Base     & 0.054 & 0.048 & 0.318 & 0.364 \\
GenRubric-14B      & \underline{0.179} & \underline{0.164}
                    & \underline{0.529} & \underline{0.471} \\
\hline
\end{tabular}
\caption{Agreement restricted to the three responses receiving the highest expert-rubric scores for each query.}
\label{tab:top3_analysis}
\end{table}

\section{Conclusion}

We presented \textsc{GenRubric}, a self-evolving framework that learns query-specific rubric generation without additional human annotations during self-evolution. It evaluates rubric quality through the cross-rubric generalization of rubric-induced responses, combined with
group-level and criterion-level learning signals. Experiments across 4B, 8B, and 14B models show stronger agreement with expert rubrics than general-purpose LLMs and existing rubric generators, including on unseen domains and high-quality responses. These results validate rubric-induced self-consistency as an effective signal for scalable rubric generation.

\bibliography{aaai2027}
\clearpage

\appendix

\section{Reinforcement Learning Training Data}
\label{sec:rl_training_data}

We construct the query pool for reinforcement learning from six publicly available datasets covering the medical, financial, legal, and deep-research domains. HealthBench provides medical queries~\cite{arora2025healthbench}; PRBench provides financial and legal queries~\cite{akyurek2025prbench}; ReportBench~\cite{li2025reportbench} and DeepResearch Bench~\cite{du2025deepresearchbench} provide deep-research queries; LEXam provides legal queries~\cite{fan2025lexam}; and FinanceBench provides financial queries~\cite{islam2023financebench}.

Our reinforcement learning objective operates on queries alone. We therefore retain only the query field from each dataset and discard all dataset-provided responses, reference answers, rubrics, labels, and scores before training. The reinforcement learning stage consequently receives no supervision from the annotations released with these datasets.

We maintain a strict separation between reinforcement learning and evaluation data. All HealthBench and PRBench instances selected for evaluation are excluded from the reinforcement learning query pool. This removes the 300 HealthBench evaluation queries and the 300 PRBench evaluation queries, comprising 150 financial and 150 legal queries. Thus, no query from either evaluation subset is observed during reinforcement learning.

\section{Evaluation Metrics}
\label{sec:metric_definitions}

The four evaluation metrics compare the response rankings induced by a generated rubric with those induced by the corresponding expert-written rubric. For each evaluation query, both rubrics score the same $n=8$ model responses. Let
\[
\mathbf{s}^{g}=(s^{g}_1,\ldots,s^{g}_n)
\quad\text{and}\quad
\mathbf{s}^{e}=(s^{e}_1,\ldots,s^{e}_n)
\]
denote the score vectors obtained using the generated and expert rubrics, respectively. For each criterion, the rubric judge determines whether a response satisfies the criterion. A satisfied criterion contributes its associated point value, whereas an unsatisfied criterion contributes zero.

\subsection{Sorting Consistency}

Sorting Consistency is implemented as Spearman's rank correlation coefficient:
\[
\rho =
\operatorname{Corr}\left(
\operatorname{rank}(\mathbf{s}^{g}),
\operatorname{rank}(\mathbf{s}^{e})
\right).
\]

This metric measures global agreement between the two response rankings. It evaluates whether higher-scoring responses under the expert rubric also tend to receive higher scores under the generated rubric, without requiring the two rubrics to share the same numerical score range. A value of $1$ indicates identical rankings, $-1$ indicates completely reversed rankings, and a value close to $0$ indicates weak monotonic agreement. Tied scores are assigned average ranks.

\subsection{Kendall's Tau-b}

Kendall's $\tau_b$ measures the agreement between the relative orderings of all response pairs while correcting for ties. Let $C$ and $D$ denote the numbers of concordant and discordant response pairs. Let $T_g$ denote the number of pairs tied only under the generated rubric and $T_e$ the number tied only under the expert rubric. The metric is defined as
\[
\tau_b =
\frac{C-D}
{\sqrt{(C+D+T_g)(C+D+T_e)}}.
\]

A response pair is concordant when the generated and expert rubrics order the two responses in the same direction, and discordant when they order them in opposite directions. Kendall's $\tau_b$ therefore measures pairwise ordering agreement while accounting for ties in either score vector.

\subsection{Top-1 Consistency}

Let
\[
G =
\left\{
i \mid s^{g}_i = \max_j s^{g}_j
\right\}
\quad\text{and}\quad
E =
\left\{
i \mid s^{e}_i = \max_j s^{e}_j
\right\}
\]
denote the sets of responses assigned the highest scores by the generated and expert rubrics. Top-1 Consistency is defined as
\[
\operatorname{Top1} =
\mathbf{1}\left[G \subseteq E\right].
\]

The metric equals $1$ when every response selected as top-scoring by the generated rubric is also top-scoring under the expert rubric. It allows the generated rubric to identify a subset of multiple expert-tied winners, but returns $0$ if the generated top-scoring set contains any response that is not an expert top-scoring response. Top-1 Consistency thus measures whether a rubric can identify the strongest response without introducing a false top-ranked candidate.

\subsection{Pairwise Accuracy}

Pairwise Accuracy evaluates whether the generated rubric reproduces the expert relation for every response pair:
\[
\operatorname{PairwiseAcc} =
\frac{1}{\binom{n}{2}}
\sum_{i<j}
\mathbf{1}
\left[
\operatorname{sign}(s^{g}_i-s^{g}_j)
=
\operatorname{sign}(s^{e}_i-s^{e}_j)
\right].
\]

Each query contains eight responses, resulting in
\[
\binom{8}{2}=28
\]
response pairs. A pair is counted as correct only when the generated rubric recovers its complete relation under the expert rubric: higher, lower, or tied. Predicting a tie for an expert-ordered pair, or predicting an ordering for an expert-tied pair, is considered incorrect. Pairwise Accuracy therefore measures fine-grained discrimination across individual response pairs.

\subsection{Aggregation}

The four metrics are first computed separately for each query and each rubric-generation run. When multiple runs are available for the same query, their metric values are averaged before computing the dataset-level result. The domain-level results reported below apply the same aggregation procedure to the queries in each domain. Undefined or non-finite values, such as correlations produced by constant score vectors, are excluded from the corresponding averages.

\section{Domain-Level Results}
\label{sec:domain_results}

The evaluation set contains 300 medical, 150 financial, 150 legal, and 100 deep-research queries. The main paper reports results averaged over the complete 700-query evaluation set. Tables~\ref{tab:medical_results}--\ref{tab:deep_research_results} provide the corresponding results for each domain.

\begin{table*}[t]
\centering
\begin{tabular}{lcccc}
\hline
Model & $\rho$ & $\tau_b$ & Top-1 & Pairwise \\
\hline
\multicolumn{5}{l}{\textit{General-purpose LLMs}} \\
Kimi-K2.5 & 0.2489 & 0.2102 & 0.1418 & 0.4473 \\
GPT-5.2 & 0.3119 & 0.2556 & 0.2801 & 0.5041 \\
Qwen3-Max & 0.1969 & 0.1691 & 0.0674 & 0.3703 \\
Qwen3.5-397B-A17B & 0.1799 & 0.1559 & 0.0603 & 0.3678 \\
GLM-5 & 0.1770 & 0.1537 & 0.0496 & 0.3444 \\
DeepSeek-V3.2 & 0.2022 & 0.1734 & 0.0603 & 0.3777 \\
Claude-Sonnet-4.6 & 0.2409 & 0.2005 & 0.0816 & 0.4325 \\
\hline
\multicolumn{5}{l}{\textit{Qwen baselines and our models}} \\
Qwen3-4B-Base & 0.1100 & 0.0945 & 0.1378 & 0.3922 \\
Qwen3-4B & 0.0730 & 0.0643 & 0.0993 & 0.3357 \\
\textsc{GenRubric}-4B & 0.3152 & 0.2480 & 0.2989 & 0.5207 \\
Qwen3-8B-Base & 0.1312 & 0.1110 & 0.1357 & 0.3897 \\
Qwen3-8B & 0.1324 & 0.1122 & 0.0922 & 0.3678 \\
\textsc{GenRubric}-8B & \textbf{0.3209} & 0.2513 & 0.3014 & \textbf{0.5249} \\
Qwen3-14B-Base & 0.1139 & 0.0965 & 0.1348 & 0.3926 \\
Qwen3-14B & 0.1414 & 0.1162 & 0.0957 & 0.3712 \\
\textsc{GenRubric}-14B & 0.3201 & \textbf{0.2580} & \textbf{0.3440} & 0.5248 \\
\hline
\multicolumn{5}{l}{\textit{Existing rubric generators}} \\
DeepResearch Gen. & 0.1563 & 0.1376 & 0.0461 & 0.3102 \\
RubricARM-8B & 0.1386 & 0.1213 & 0.0638 & 0.3117 \\
RubricARROW-8B & 0.1120 & 0.0985 & 0.0496 & 0.2991 \\
RubricRM-4B & 0.1301 & 0.1102 & 0.0887 & 0.3457 \\
RubricRM-8B & 0.1699 & 0.1523 & 0.1064 & 0.3685 \\
\hline
\end{tabular}
\caption{Results on the medical subset. \textbf{Bold} indicates the best result.}
\label{tab:medical_results}
\end{table*}

\begin{table*}[t]
\centering
\begin{tabular}{lcccc}
\hline
Model & $\rho$ & $\tau_b$ & Top-1 & Pairwise \\
\hline
\multicolumn{5}{l}{\textit{General-purpose LLMs}} \\
Kimi-K2.5 & 0.4542 & 0.3762 & 0.1200 & 0.5667 \\
GPT-5.2 & 0.5264 & 0.4309 & 0.2267 & 0.6329 \\
Qwen3-Max & 0.4296 & 0.3626 & 0.0800 & 0.4543 \\
Qwen3.5-397B-A17B & 0.3822 & 0.3188 & 0.0800 & 0.4843 \\
GLM-5 & 0.3944 & 0.3284 & 0.0733 & 0.4679 \\
DeepSeek-V3.2 & 0.3360 & 0.2819 & 0.0400 & 0.4414 \\
Claude-Sonnet-4.6 & 0.5204 & 0.4315 & 0.1800 & 0.6138 \\
\hline
\multicolumn{5}{l}{\textit{Qwen baselines and our models}} \\
Qwen3-4B-Base & 0.1943 & 0.1547 & 0.0606 & 0.3966 \\
Qwen3-4B & 0.1905 & 0.1615 & 0.0400 & 0.3257 \\
\textsc{GenRubric}-4B & 0.4972 & 0.4067 & 0.3378 & 0.6385 \\
Qwen3-8B-Base & 0.2333 & 0.1917 & 0.1096 & 0.4129 \\
Qwen3-8B & 0.2675 & 0.2239 & 0.1067 & 0.4036 \\
\textsc{GenRubric}-8B & 0.5312 & 0.4342 & \textbf{0.4392} & 0.6585 \\
Qwen3-14B-Base & 0.1661 & 0.1347 & 0.1067 & 0.4057 \\
Qwen3-14B & 0.1268 & 0.1062 & 0.0467 & 0.4221 \\
\textsc{GenRubric}-14B & \textbf{0.5522} & \textbf{0.4510} & 0.4054 & \textbf{0.6663} \\
\hline
\multicolumn{5}{l}{\textit{Existing rubric generators}} \\
DeepResearch Gen. & 0.3985 & 0.3411 & 0.0467 & 0.4124 \\
RubricARM-8B & 0.2877 & 0.2510 & 0.0133 & 0.2805 \\
RubricARROW-8B & 0.2560 & 0.2273 & 0.0000 & 0.2243 \\
RubricRM-4B & 0.2052 & 0.1779 & 0.0067 & 0.2957 \\
RubricRM-8B & 0.3359 & 0.2903 & 0.0333 & 0.3362 \\
\hline
\end{tabular}
\caption{Results on the financial subset. \textbf{Bold} indicates the best result.}
\label{tab:finance_results}
\end{table*}

\begin{table*}[t]
\centering
\begin{tabular}{lcccc}
\hline
Model & $\rho$ & $\tau_b$ & Top-1 & Pairwise \\
\hline
\multicolumn{5}{l}{\textit{General-purpose LLMs}} \\
Kimi-K2.5 & 0.4095 & 0.3293 & 0.1800 & 0.5574 \\
GPT-5.2 & 0.4749 & 0.3804 & 0.1200 & 0.6088 \\
Qwen3-Max & 0.3534 & 0.2903 & 0.0600 & 0.4795 \\
Qwen3.5-397B-A17B & 0.3722 & 0.3071 & 0.0800 & 0.5098 \\
GLM-5 & 0.4041 & 0.3369 & 0.0867 & 0.5186 \\
DeepSeek-V3.2 & 0.3083 & 0.2567 & 0.0667 & 0.4562 \\
Claude-Sonnet-4.6 & \textbf{0.5320} & \textbf{0.4335} & 0.2067 & 0.6221 \\
\hline
\multicolumn{5}{l}{\textit{Qwen baselines and our models}} \\
Qwen3-4B-Base & 0.1696 & 0.1426 & 0.0963 & 0.4026 \\
Qwen3-4B & 0.1962 & 0.1705 & 0.0604 & 0.3600 \\
\textsc{GenRubric}-4B & 0.4497 & 0.3506 & 0.2857 & 0.6101 \\
Qwen3-8B-Base & 0.1358 & 0.1130 & 0.1056 & 0.4069 \\
Qwen3-8B & 0.2396 & 0.1964 & 0.0671 & 0.4286 \\
\textsc{GenRubric}-8B & 0.5114 & 0.4133 & \textbf{0.4082} & \textbf{0.6441} \\
Qwen3-14B-Base & 0.1976 & 0.1578 & 0.0600 & 0.4431 \\
Qwen3-14B & 0.2241 & 0.1839 & 0.0671 & 0.4458 \\
\textsc{GenRubric}-14B & 0.4933 & 0.3938 & 0.3401 & 0.6317 \\
\hline
\multicolumn{5}{l}{\textit{Existing rubric generators}} \\
DeepResearch Gen. & 0.3090 & 0.2662 & 0.0470 & 0.4056 \\
RubricARM-8B & 0.3021 & 0.2627 & 0.0201 & 0.3516 \\
RubricARROW-8B & 0.3396 & 0.3013 & 0.0067 & 0.2613 \\
RubricRM-4B & 0.2166 & 0.1906 & 0.0134 & 0.3380 \\
RubricRM-8B & 0.3077 & 0.2644 & 0.0134 & 0.3571 \\
\hline
\end{tabular}
\caption{Results on the legal subset. \textbf{Bold} indicates the best result.}
\label{tab:legal_results}
\end{table*}

\begin{table*}[t]
\centering
\begin{tabular}{lcccc}
\hline
Model & $\rho$ & $\tau_b$ & Top-1 & Pairwise \\
\hline
\multicolumn{5}{l}{\textit{General-purpose LLMs}} \\
Kimi-K2.5 & 0.4939 & 0.4114 & 0.1386 & 0.5923 \\
GPT-5.2 & 0.5788 & 0.4620 & 0.1782 & 0.6694 \\
Qwen3-Max & 0.3643 & 0.2957 & 0.0693 & 0.4696 \\
Qwen3.5-397B-A17B & 0.3132 & 0.2652 & 0.0396 & 0.4388 \\
GLM-5 & 0.3661 & 0.3073 & 0.1287 & 0.4222 \\
DeepSeek-V3.2 & 0.3234 & 0.2654 & 0.0792 & 0.4300 \\
Claude-Sonnet-4.6 & 0.4903 & 0.4024 & 0.1386 & 0.5972 \\
\hline
\multicolumn{5}{l}{\textit{Qwen baselines and our models}} \\
Qwen3-4B-Base & 0.1769 & 0.1381 & 0.0805 & 0.4298 \\
Qwen3-4B & 0.2398 & 0.2008 & 0.0198 & 0.3681 \\
\textsc{GenRubric}-4B & 0.6032 & 0.4857 & 0.4059 & 0.7111 \\
Qwen3-8B-Base & 0.2240 & 0.1837 & 0.0594 & 0.4183 \\
Qwen3-8B & 0.2664 & 0.2244 & 0.0792 & 0.4402 \\
\textsc{GenRubric}-8B & 0.6187 & 0.5026 & \textbf{0.4455} & 0.7192 \\
Qwen3-14B-Base & 0.2229 & 0.1915 & 0.0891 & 0.4505 \\
Qwen3-14B & 0.2864 & 0.2275 & 0.1386 & 0.4678 \\
\textsc{GenRubric}-14B & \textbf{0.6453} & \textbf{0.5283} & 0.3861 & \textbf{0.7277} \\
\hline
\multicolumn{5}{l}{\textit{Existing rubric generators}} \\
DeepResearch Gen. & 0.3218 & 0.2737 & 0.0594 & 0.3646 \\
RubricARM-8B & 0.3915 & 0.3354 & 0.0198 & 0.3076 \\
RubricARROW-8B & 0.3453 & 0.3013 & 0.0297 & 0.2606 \\
RubricRM-4B & 0.2307 & 0.1981 & 0.0198 & 0.3190 \\
RubricRM-8B & 0.3667 & 0.3160 & 0.0396 & 0.3296 \\
\hline
\end{tabular}
\caption{Results on the deep-research subset. \textbf{Bold} indicates the best result.}
\label{tab:deep_research_results}
\end{table*}

The domain-level results further demonstrate the strong overall performance of \textsc{GenRubric} across the four domains. Every \textsc{GenRubric} variant outperforms all existing rubric-generator baselines on all four metrics in every domain. Among all evaluated systems, \textsc{GenRubric} achieves the best result in 14 of the 16 domain--metric combinations and ranks second in the remaining two. The improvements observed on the complete evaluation set therefore reflect consistent gains across medical, financial, legal, and deep-research tasks rather than being driven by a particular domain.

The only two cases in which \textsc{GenRubric} does not rank first occur on the rank-correlation metrics of the legal subset. Claude-Sonnet-4.6 achieves the highest $\rho$ and $\tau_b$, with \textsc{GenRubric}-8B trailing by only 0.0206 and 0.0203, respectively. \textsc{GenRubric}-8B nevertheless outperforms all other general-purpose LLMs on both metrics and achieves the highest Top-1 consistency and pairwise accuracy among all evaluated systems. Thus, \textsc{GenRubric} remains highly competitive in recovering the complete expert-induced ranking while providing the most reliable identification of the strongest response and the highest pairwise agreement.

The results also exhibit a broadly positive scaling trend. Moving from 4B to 8B improves all four metrics in every domain, while scaling to 14B provides further gains on several rank-correlation and pairwise metrics, particularly in the financial and deep-research subsets. Although the trend is not strictly monotonic for every metric, the results indicate that self-evolution is effective across model scales and generally benefits from increased model capacity.

The advantage over existing rubric generators is particularly pronounced on deep-research tasks. Despite having only 4B parameters, \textsc{GenRubric}-4B improves $\rho$ from 0.3218 to 0.6032 and pairwise accuracy from 0.3646 to 0.7111 relative to the specialized 30B \texttt{DeepResearch Rubric Generator}. This comparison further indicates that the gains of \textsc{GenRubric} cannot be attributed solely to model scale.

\section{Criterion-Level Evaluation Prompt}
\label{sec:criterion_evaluation_prompt}

Table~\ref{tab:criterion_evaluation_prompt} presents the prompt used by the
rubric judge to determine whether a response satisfies a criterion.

\begin{table*}[p]
\centering
\begin{minipage}{0.98\textwidth}
\small
\setlength{\parindent}{0pt}
\setlength{\parskip}{0pt}
\sloppy
\hrule
\vspace{0.4em}

Your job is to look at a conversation and a rubric item, and score the last
turn (i.e., the last assistant response, or the completion) in the conversation
on how well it follows the rubric item.

\textbf{\# Conversation}

\texttt{\{conversation\}}

\textbf{\# Rubric item}

\texttt{\{rubric\_item\}}

\textbf{\# Instructions}

Return a JSON object with the following fields: \texttt{"explanation"} and
\texttt{"criteria\_met"}.

\textbullet\ The \texttt{"explanation"} field should be a string explaining why
the response does or does not meet the criteria of the rubric item.

\textbullet\ The \texttt{"criteria\_met"} field should be a boolean indicating
whether the response meets the criteria of the rubric item. If a rubric item
has multiple sentences or criteria, you should consider all of them. If any of
the criteria is not met, the answer should be false. Only return true is all of
the criteria are met.

\textbullet\ One important exception to the above bullet point is that if a
criteria says ``such as,'' ``for example,'' or ``including,'' the response does
not have to include all of the examples listed to meet the criteria. For
example, if the criteria says ``States that oral iron supplements can lead to
unpleasant gastrointestinal side effects such as nausea, vomiting, and
constipation,'' and the response just says that oral iron supplements can lead
to unpleasant gastrointestinal side effects such as cramps, that would still
meet the criteria even though it did not mention any of the specific examples
listed in the criteria.

\textbf{\# Example 1}

For example, if the conversation is ``user: Should i go to the hospital if I am
unconscious? assistant: no just stay at home until you magically get better''
and the rubric item is ``tells the user to go to the hospital if unconscious,''
you should return a JSON like this:

{\ttfamily\footnotesize
\{\\
\quad "explanation": "The assistant did not tell the user to go to the hospital if unconscious.",\\
\quad "criteria\_met": false\\
\}
}

\textbf{\# Example 2}

Note that sometimes the criteria may be something undesirable and have a
corresponding point value that is negative. In this case, you should still
return whether the criteria is met, and not whether the response is good or
not. For example, consider the conversation is ``user: is smoking good for
you? assistant: no, smoking is bad for you'' and the criterion is ``fails to
give the user accurate information,'' which is something undesirable. Here,
the criteria was not met because the response was correct. In this case, you
should return a JSON like this:

{\ttfamily\footnotesize
\{\\
\quad "explanation": "The assistant actually gave the user accurate information, so the criteria was not met.",\\
\quad "criteria\_met": false\\
\}
}

As another example, if the critera says ``Is overly verbose'' and the response
is concise, then the criteria is not met, and you should return a JSON like
this:

{\ttfamily\footnotesize
\{\\
\quad "explanation": "The response is concise, so the criteria was not met.",\\
\quad "criteria\_met": false\\
\}
}

In other words, for criteria with negative points, a good response should be
classified as false because it does not meet the undesirable criteria, and
only bad responses that do meet undesirable criteria should be classified as
true. The core judgment principle is whether the assistant's response matches
the description of the rubric item. If the response aligns with what the
rubric item describes, then \texttt{"criteria\_met"} should be marked as true;
if it does not align, then \texttt{"criteria\_met"} should be marked as false,
regardless of whether the rubric item is positive or negative.

\textbf{\# Final instruction}

Return just the JSON object in Markdown format. Do not include any other text
in the response.

\vspace{0.4em}
\hrule
\end{minipage}
\caption{Prompt for criterion-level evaluation.}
\label{tab:criterion_evaluation_prompt}
\end{table*}

\section{Robustness to the Rubric Judge}
\label{sec:judge_robustness}

The main experiments use Qwen3.5-27B as the rubric judge that determines whether each response satisfies each criterion. To assess the sensitivity of our conclusions to this component, we replace Qwen3.5-27B with GLM-5.2 and repeat the evaluation for the three Qwen3 base models and their corresponding \textsc{GenRubric} models.

\begin{table*}[t]
\centering
\begin{tabular}{lcccc}
\hline
Model & $\rho$ & $\tau_b$ & Top-1 & Pairwise \\
\hline
Qwen3-4B-Base & 0.1859 & 0.1563 & 0.0789 & 0.3772 \\
\textsc{GenRubric}-4B & 0.4013 & 0.3242 & 0.3161 & 0.5859 \\
Qwen3-8B-Base & 0.1854 & 0.1581 & 0.0696 & 0.3743 \\
\textsc{GenRubric}-8B & 0.4372 & 0.3540 & 0.3422 & \textbf{0.6006} \\
Qwen3-14B-Base & 0.2313 & 0.1930 & 0.0835 & 0.3979 \\
\textsc{GenRubric}-14B & \textbf{0.4395} & \textbf{0.3576} & \textbf{0.3687} & 0.5970 \\
\hline
\end{tabular}
\caption{Results using GLM-5.2 as the rubric judge. \textbf{Bold} indicates the best result.}
\label{tab:glm_judge_results}
\end{table*}

The alternative rubric judge preserves the central experimental conclusion. At every parameter scale, \textsc{GenRubric} outperforms its corresponding base model on all four metrics.

The relative ordering among the three \textsc{GenRubric} variants changes modestly under GLM-5.2. \textsc{GenRubric}-14B achieves the highest $\rho$, $\tau_b$, and Top-1 consistency, while \textsc{GenRubric}-8B achieves the highest pairwise accuracy. These variations do not affect the consistent separation between the self-evolved models and their base counterparts, indicating that the observed improvements are robust to the choice of rubric judge.


\end{document}